\documentclass[11pt]{article}

\usepackage[preprint]{acl}

\usepackage{times}
\usepackage{latexsym}

\usepackage[T1]{fontenc}
\usepackage[utf8]{inputenc}

\usepackage{microtype}

\usepackage{inconsolata}

\usepackage{graphicx}

\usepackage{booktabs}   % 三线表
\usepackage{multirow}   % 跨行单元格
\usepackage{makecell}   % 单元格内换行与样式
\usepackage{tabularx}   % 自动伸缩等宽列
\usepackage[table]{xcolor} % 颜色支持（集成 colortbl 功能）

\usepackage{enumitem}   % 列表格式控制
\usepackage{mdframed}   % 边框容器
\usepackage{amsmath}
 
\newcolumntype{Y}{>{\centering\arraybackslash}X}

\definecolor{labelbg}{RGB}{235, 248, 235} 
\definecolor{keywordblue}{RGB}{0, 0, 200}
\definecolor{commentpink}{RGB}{230, 0, 100}
\definecolor{graybg}{gray}{0.9}
\definecolor{gainblue}{RGB}{60, 100, 255}

\newcommand{\gain}[1]{\textcolor{gainblue}{(+#1)}}
\title{EMR: Self-Evolving Medical Multi-Agent System via \\Experience Mining and Reuse}

\author{  
    Dongsheng Shi\textsuperscript{\rm 1}, Yue Li\textsuperscript{\rm 1}, Xin Yi\textsuperscript{\rm 1}, \textbf{Linlin Wang$^{1,2}$\thanks{Corresponding Author.}} \\  
    $^{1}$East China Normal University\\ 
    $^{2}$City University of Hong Kong\\ 
    \texttt{\{dongsheng, yue\_li, xinyi\}@stu.ecnu.edu.cn,} \\ \texttt{llwang@cs.ecnu.edu.cn} \\  
}

\begin{document}
\maketitle
\begin{abstract}
Large language model (LLM) driven multi-agent systems have shown promise in complex clinical reasoning, yet existing approaches rely on static strategies and lack persistent clinical memory, preventing self-evolving from prior diagnostic successes and failures.
We present EMR, a self-evolving medical multi-agent system via \underline{\textbf{E}}xperience \underline{\textbf{M}}ining and \underline{\textbf{R}}euse. EMR introduces a hierarchical clinical experience library that organizes accumulated knowledge into three levels: clinical principles, diagnostic patterns, and representative cases. During inference, EMR emulates multidisciplinary consultation: a planner agent coordinates domain-specific department agents for specialized reasoning, while a summary agent synthesizes their analyses into a final decision. Critically, EMR automatically extracts correct diagnostic insights and failure-related warnings from multi-agent reasoning trajectories, incrementally updating the experience library to guide future cases.
Experiments on medical reasoning benchmarks demonstrate that EMR consistently outperforms state-of-the-art medical multi-agent baselines. Further analysis reveals that the hierarchical experience enables cross-specialty generalization and transfer across diverse LLM backbones, offering a scalable and interpretable pathway toward continually evolving intelligent medical reasoning.
% Project Page: \url{https://anonymous.4open.science/r/EMR_dev-A2C1/}
\end{abstract}

\section{Introduction}

\begin{figure}[h]
    \centering
    \includegraphics[width=\linewidth]{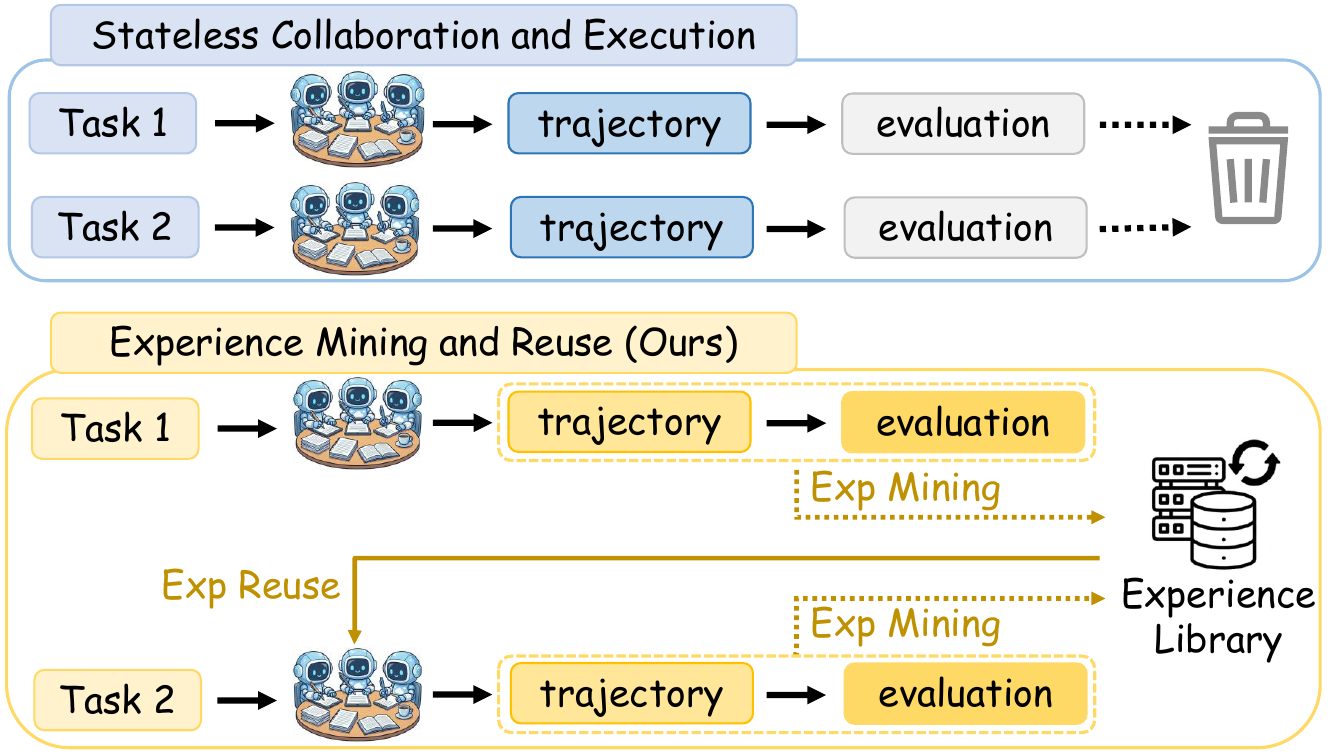}
    \caption{Illustration of the experience mining and reuse mechanism in EMR versus traditional instance-independent reasoning.}
    \label{fig:intro}
\end{figure}

% 背景
% 大型语言模型（LLM）推动了多智能体系统的发展
% 医疗多智能体
Recent advances in large language models (LLMs) have substantially improved their capabilities across a wide range of tasks, particularly in reasoning, planning, and tool use \cite{yi2025unified, li2025hierarchical, chen2026reinforcement, zhang2026tools}. This progress has enabled the rise of autonomous agents and subsequently stimulated the development of multi-agent systems, where multiple agents work together to address complex tasks \cite{autogpt2023, xagent2023, hong2024metagpt, chen2024reconcile, shi2026surgent}.
% Large language models (LLMs) have driven the development of multi-agent systems, in which multiple autonomous agents work together to solve complex tasks \cite{autogpt2023, xagent2023, hong2024metagpt, chen2024reconcile, shi2026surgent}. 
In medical settings, such complexity often arises from multi-symptom presentations, differential diagnosis, and the need for cross-department clinical consultation \cite{202510.1448, shi2026benchmarking}.
In this context, multi-agent systems have been increasingly explored in the medical domain, as they can decompose complex medical problems into coordinated reasoning processes across multiple specialized agents \cite{tang2024medagents, kim2024mdagents}.

% 问题
However, existing medical multi-agent systems suffer from two key limitations.
(1) Most current approaches are \textbf{static}, relying on fixed reasoning strategies, which often lead to recurring diagnostic errors when facing complex or previously unseen clinical cases \cite{202510.1448, MIAO2026106136, 202512.2602}.
(2) As illustrated in Figure \ref{fig:intro}, medical multi-agent systems typically treat each task as an \textbf{isolated} instance, failing to learn from past successful clinical experiences or to avoid previously encountered diagnostic mistakes \cite{flesch2018comparing}.
This lack of experience accumulation fundamentally limits their ability to improve robustness and reliability in medical reasoning.

% 我们认为缺失的要素是显式且持久的经验抽象。多智能体系统不应孤立地演化提示或工作流，而应从自身的推理轨迹中提取可重用的知识，并以可迁移、可扩展且与模型无关的形式存储。这种经验不仅应捕捉成功的推理策略，还应捕捉系统性的失败模式，从而使智能体能够避免重蹈覆辙。在临床实践中，这种经验自然对应于积累的诊断指南、常用的推理启发式方法以及从以往成功和失败中汲取的代表性病例。
We argue that the missing component is an explicit and persistent experience abstraction. Rather than evolving prompts \cite{yuksekgonul2024textgrad, zhang2025agentic} or workflows \cite{lin2025se, zhang2025multi, shang2024agentsquare} in isolation, multi-agent systems should extract reusable knowledge from their own reasoning trajectories and store it in a form that is transferable, scalable, and model-agnostic. Such experience should capture not only successful reasoning strategies but also systematic failure modes, enabling agents to avoid repeating past mistakes \cite{cai2025flex, wu2025evolver}. In clinical practice, such experience naturally corresponds to accumulated diagnostic guidelines, common reasoning heuristics, and representative patient cases learned from prior successes and failures.

% 为此，我们提出了一种名为EMR的医学多智能体系统，该系统具有经验积累能力，能够通过经验挖掘和重用实现持续学习。EMR的多智能体架构旨在模拟多学科临床决策工作流程。规划智能体负责分配相关的医疗科室，科室智能体生成特定领域的推理轨迹，而总结智能体则整合跨科室分析结果，最终生成决策。在多智能体协作过程中，EMR能够自动从智能体轨迹中挖掘黄金经验和警告经验，并逐步更新其经验库。
To this end, we propose EMR, a medical multi-agent system with experience, which enables self-evolving through experience mining and reuse. The multi-agent architecture of EMR is designed to mirror multidisciplinary clinical decision-making workflows. Specifically, a planner agent assigns relevant medical departments, department agents generate domain-specific reasoning trajectories, and a summary agent integrates cross-department analyses to produce final decisions. During the multi-agent collaboration process, EMR automatically mines both golden and warning experiences from agent trajectories and incrementally updates its experience library.

\begin{figure*}[ht]
  \centering
  \includegraphics[width=0.9\textwidth]{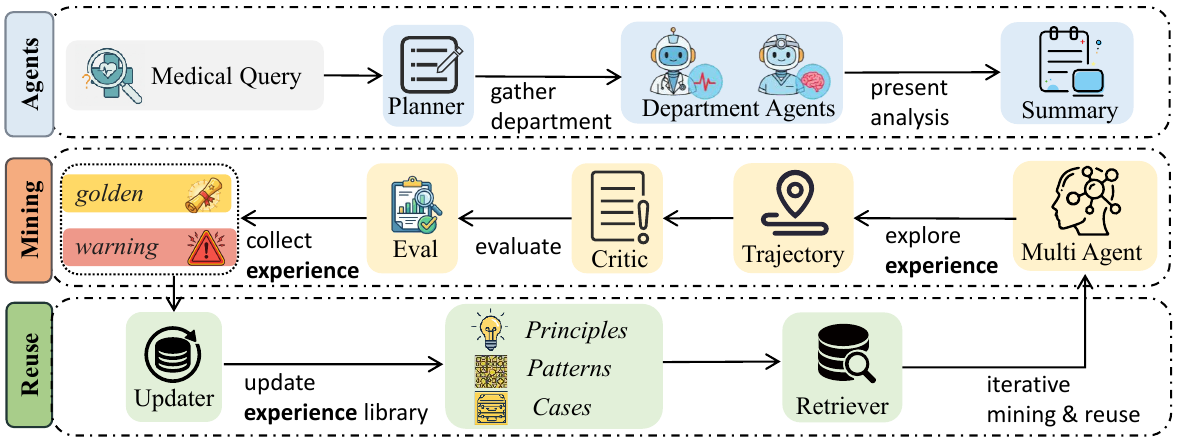}
  \caption{Our proposed EMR diagram.
Agents correspond to Section~\ref{subsection:multi-agent} and represent the multi-agent collaborative system.
Mining corresponds to Section~\ref{subsection:mining} and focuses on analyzing multi-agent collaboration trajectories to mine experiential knowledge.
Reuse corresponds to Section~\ref{subsection:reuse}, where extracted experiences are integrated into an experience library and reused to guide subsequent tasks, enabling iterative mining and reuse.}
  \label{fig:framework}
\end{figure*}

% 至关重要的是，EMR 将经验组织成一个三级层级结构——原则、模式和案例——分别对应于临床指南、常用诊断启发式方法和代表性患者案例。这种层级设计使得高级医学推理策略能够跨任务泛化，同时在需要时保留细粒度的临床细节。在推理过程中，智能体会重用相关经验来指导推理，从而使系统能够在不修改模型参数的情况下持续提升临床推理性能。
Crucially, EMR organizes experience into a three-level hierarchy—principles, patterns, and cases—corresponding to clinical guidelines, common diagnostic heuristics, and representative patient cases, respectively. This hierarchical design allows high-level medical reasoning strategies to generalize across tasks while preserving fine-grained clinical details when needed. During inference, agents reuse relevant experiences to guide reasoning, enabling the system to continuously improve clinical reasoning performance without modifying model parameters.

% 我们在多个医学问答基准测试中评估了 EMR。实验结果表明，EMR 始终优于静态多智能体基线，平均准确率比 MDAgents 高出 3.3%、2.0% 和 1.5%。
% 详细分析表明，经验在不同的主干LLM和数据集之间具有可迁移性。
% 这些发现表明，显式经验建模为持续演进的医学多智能体系统提供了一条切实可行且可扩展的途径。
We evaluate EMR on multiple medical question answering benchmarks. Experimental results show that EMR consistently outperforms static multi-agent baselines, outperforming MDAgents \cite{kim2024mdagents} by 3.3\%, 2.0\%, and 1.5\% in average accuracy.
Detailed analysis indicates that experiences exhibit transferability across both different backbone LLMs and datasets.
These findings indicate that explicit experience modeling provides a practical and scalable path toward continuously evolving medical multi-agent systems.

The main contributions are summarized as follows:
\begin{itemize}
    % 我们提出了 EMR，这是一个医疗多智能体系统，它通过经验挖掘和在分层经验库中重用来支持持续学习。
    \item We propose EMR, a medical multi-agent system that supports self-evolving through experience mining and reuse within a hierarchical experience library.

    % 我们通过实证研究证明了 EMR 在医学问答基准测试中的有效性，突出了不同经验类型的互补作用。
    \item We empirically demonstrate the effectiveness of EMR on medical question answering benchmarks, highlighting the complementary roles of different experience types.

    % 我们证明，从不同的 LLM 或数据集中挖掘出的经验是可转移的，并且仍然是人类可读的，为改进多智能体推理策略提供了可操作的指导。
    \item We show that experiences mined from different LLMs or datasets are transferable and remain human-readable, providing actionable guidance for improving multi-agent reasoning strategies.
\end{itemize}

\section{Related Work}
\subsection{Multi-Agent Collaboration in Medical}

Recent work has explored multi-agent collaboration as a promising paradigm for improving LLM performance in medical tasks. AI Hospital \cite{fan2025ai} and Self-Evolving Multi-Agent Simulations \cite{almansoori2025self} construct realistic clinical environments where multiple medical agents interact, adapt, and evolve, enabling systematic evaluation of LLMs under complex clinical workflows. For medical reasoning and decision-making, MedAgents \cite{tang2024medagents} and MDAgents \cite{kim2024mdagents} show that collaborative LLMs with role specialization, adaptive coordination, and consensus mechanisms can significantly enhance single-LLM reasoning and clinical decision quality, while ReConcile \cite{chen2024reconcile} further demonstrates the benefits of consensus among diverse LLMs for robust reasoning. Beyond reasoning, ColaCare \cite{wang2025colacare} applies LLM-driven multi-agent collaboration to electronic health record modeling, and nomadic multi-agent systems \cite{dhasarathan2024nomadic} have been used to quantify and protect privacy in dynamic e-healthcare settings. Overall, these studies highlight the effectiveness of multi-agent collaboration in advancing medical reasoning, simulation, and clinical data modeling.

\subsection{Experience-Driven Agent Evolution}
Recently, experience-driven evolution has emerged as a promising approach for enabling LLM agents to accumulate and reuse past interaction knowledge.
ExpeL~\cite{zhao2024expel} stores raw trajectories without structured abstraction, while ReasoningBank~\cite{ouyang2025reasoningbank} generates parallel trajectories under the same model, and G-Memory~\cite{zhang2026g} encodes experiences into complex graph structures.
In the medical domain, MDTeamGPT~\cite{chen2025mdteamgpt} accumulates consultation experiences as flat summaries for specific agents, whereas EvolveR~\cite{wu2025evolver} employs GRPO to enhance experience utilization but requires additional computation and parameter updates.
Overall, these methods emphasize experience storage or retrieval rather than systematic abstraction and hierarchical reuse within collaborative multi-agent systems, leaving open how accumulated experience can be structured and transferred across diverse agents to support continuous medical reasoning improvement—an issue we address in this work.

\begin{figure}
    \centering
    \includegraphics[width=1\linewidth]{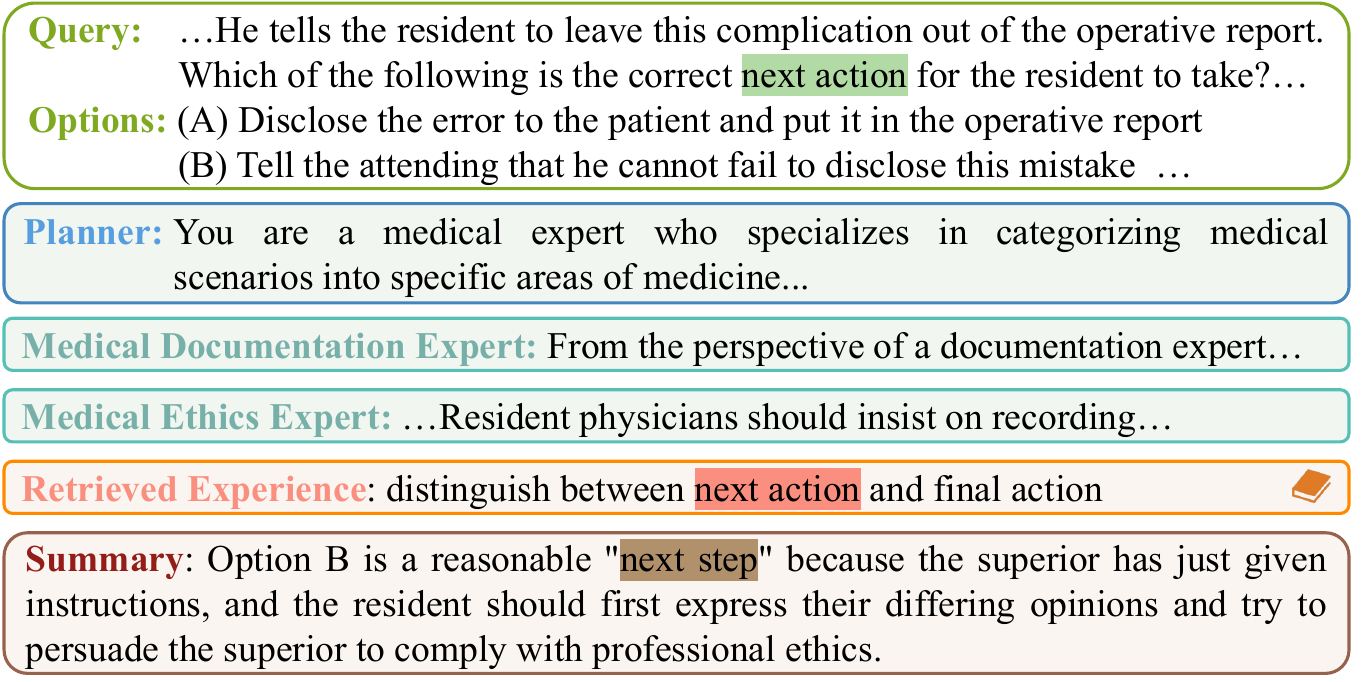}
    \caption{An illustration of EMR reasoning process.}
    \label{fig:case}
\end{figure}

\section{Method}
% 本节介绍 EMR 的三个核心组件：（i）多智能体医疗系统，（ii）经验挖掘，（iii）经验重用。
In this section, we present the components of EMR. As illustrated in Figure \ref{fig:case} and Figure \ref{fig:framework}, EMR integrates a structured multi-agent collaboration system with an explicit experience library, enabling agents to evolve without updating model parameters. The pipeline comprises three core components: (i) a multi-agent medical system (Section \ref{subsection:multi-agent}), (ii) experience mining from collaboration trajectories (Section \ref{subsection:mining}), and (iii) experience reuse through hierarchical update and retrieval (Section \ref{subsection:reuse}).

\subsection{Multi-Agent Medical System}
\label{subsection:multi-agent}
% 给定医学问题 $q$ 和选项集 $\mathcal{O}$，EMR 通过三类智能体模拟多科室会诊。
Given a medical multiple-choice question $q$ and options
$
\mathcal{O} = \{o_1, o_2, \dots, o_K\},
$
EMR simulates a multi-department clinical consultation through three types of agents: a planner agent (\ref{subsubsection:planner}), multiple department agents (\ref{subsubsection:department}), and a summary agent (\ref{subsubsection:summary}).

\subsubsection{Planner Agent}
\label{subsubsection:planner}
% 医学问题涉及多个临床领域，不同选项需从不同诊断角度推理。缺乏领域感知会导致冗余或无关探索，因此引入规划智能体显式构建推理空间。
Medical questions often span multiple clinical domains, with different options requiring reasoning from distinct diagnostic perspectives \cite{heist2014exploring, stringer2021examining}. Without explicit domain awareness, downstream agents tend toward redundant or irrelevant exploration \cite{zhang2025d3mas, aryal2024leveraging}. To address this, we introduce a planner agent that structures the reasoning space by identifying relevant medical domains at both the question and option levels.

Specifically, the planner assigns question-level domains
$
\mathcal{D}_q = \{d^{(q)}_1, \dots, d^{(q)}_{M}\},
$
capturing the core clinical specialties, and option-level domains
$
\mathcal{D}_o = \{d^{(o)}_1, \dots, d^{(o)}_{N}\}
$
for evaluating options from diverse perspectives. Incorporating retrieved prior experience $\mathcal{E}_r$ (Section~\ref{subsubsection:retrieval}), the planner leverages accumulated patterns to improve domain assignment and avoid previously observed errors. Formally:
\begin{equation}
\begin{aligned}
\mathcal{D}_q &= \mathrm{LLM}(q, \mathcal{E}_r \mid r_p, \pi_q), \\
\mathcal{D}_o &= \mathrm{LLM}(q, \mathcal{O}, \mathcal{E}_r \mid r_p, \pi_o),
\end{aligned}
\end{equation}
where $r_p$ denotes the planner role, $\pi_q$ and $\pi_o$ are prompts for question-level and option-level domain identification, respectively, both incorporating domain taxonomies and task-specific constraints.

% 规划智能体将推理空间分解为领域子问题，使下游部门智能体聚焦临床相关知识，为协调协作奠定基础。
By decomposing the reasoning space into domain-specific subproblems, the planner enables downstream department agents to focus on clinically relevant knowledge, laying the groundwork for coordinated collaboration.

\subsubsection{Department Agents}
\label{subsubsection:department}
% 准确的医疗决策需要基于临床知识和经验的深入领域分析。部门智能体模拟特定临床领域的医学专家。
While the planner structures the reasoning space, accurate medical decision-making requires in-depth, domain-specific analysis grounded in clinical knowledge and prior experience. We introduce department agents, each emulating a medical expert specialized in a particular clinical domain.

For each domain $d \in \mathcal{D}_q \cup \mathcal{D}_o$, a department agent performs focused reasoning conditioned on the question, candidate options, and retrieved experience:
\begin{equation}
a_d = \mathrm{LLM}(q, \mathcal{O}, d, \mathcal{E}_r \mid r_d, \pi_d),
\end{equation}
where $r_d$ denotes the domain expert role, and $\pi_d$ is a domain-specific reasoning prompt guiding the agent to incorporate both medical knowledge and historical experience. All analyses are collected as:
\begin{equation}
\mathcal{A} = \{a_d \mid d \in \mathcal{D}_q \cup \mathcal{D}_o\}.
\end{equation}

% 部门智能体提供互补视角供总结智能体整合。
By decomposing medical reasoning into parallel, domain-focused analyses enriched with retrieved experience, department agents provide diverse yet complementary perspectives for the summary agent to integrate.

\subsubsection{Summary Agent}
\label{subsubsection:summary}
% 有效医疗决策需将可能互补或冲突的跨领域观点整合为全局判断。总结智能体充当协调顾问。
While department agents provide parallel, domain-specific analyses, effective medical decision-making requires reconciling potentially complementary or conflicting viewpoints into a coherent global judgment. We introduce a Summary Agent that acts as a coordinating medical consultant, integrating cross-department evidence to derive the final prediction.

Beyond simple aggregation, the summary agent performs reflective reasoning by jointly considering the department analyses and retrieved experience, enabling it to resolve contradictions and correct local errors using accumulated prior experiences. Formally:
\begin{equation}
\hat{y}, \; a_{\text{sum}} = \mathrm{LLM}(q, \mathcal{O}, \mathcal{A}, \mathcal{E}_r \mid r_s, \pi_s),
\end{equation}
where $\hat{y}$ is the predicted answer, $a_{\text{sum}}$ the synthesized analysis, $r_s$ the summary role, and $\pi_s$ an aggregation prompt.

% 总结智能体联合利用多部门分析和经验检索，产生全局一致的推理结果。
By jointly leveraging multi-department analyses and experience retrieval, the summary agent produces a globally consistent reasoning outcome that reflects both distributed expert knowledge and accumulated experiential insights.

\subsection{Experience Mining}
\label{subsection:mining}
% EMR 从协作轨迹中显式挖掘经验以实现持续学习，包括轨迹生成和经验提取。
EMR enables self-evolving by explicitly mining experiences from multi-agent collaboration trajectories. This process involves trajectory generation (\ref{subsubsection:trajectory generation}) and experience collection (\ref{subsubsection:experience collection}).

\subsubsection{Multi-Agent Collaboration Trajectory Generation}
\label{subsubsection:trajectory generation}
% EMR 将多智能体推理过程记录为结构化轨迹，以支持经验驱动的演化。每条轨迹包含完整的推理步骤。
To support experience-driven evolution, EMR records the entire multi-agent reasoning process as a structured trajectory. In medical settings, learning from experience critically depends on understanding how decisions are made, which intermediate judgments contribute to success or failure, and where errors originate.

For each sample, EMR produces a complete reasoning trajectory:
\begin{equation}
\tau = \big(q, \mathcal{O}, \mathcal{D}_q, \mathcal{D}_o, \mathcal{A}, a_{\text{sum}}, \hat{y}\big),
\end{equation}
where the planner specifies relevant domains, department agents generate domain-specific analyses, and the summary agent synthesizes cross-department reasoning into a final decision. This trajectory captures intermediate reasoning steps and their organizational structure, forming the fundamental unit for subsequent experience mining.

To distinguish effective from erroneous reasoning, each trajectory is evaluated against the ground-truth answer $y$. A trajectory is successful if $\hat{y} = y$, and failed otherwise. This binary outcome provides a supervision signal enabling EMR to extract reusable experience from both correct reasoning paths and informative failures.

\subsubsection{Experience Collection and Categorization}
\label{subsubsection:experience collection}
% 推理轨迹需提炼为紧凑、可迁移的经验表征。EMR 从轨迹中提取自然语言表达的经验条目，分为两类。
Reasoning trajectories alone do not directly constitute reusable knowledge. EMR distills trajectories into compact, transferable experience representations that abstract away case-specific details while preserving critical reasoning signals. Learning in medical reasoning requires leveraging both successful decision patterns and informative failures that reveal systematic weaknesses.

Given a completed trajectory $\tau$, EMR extracts experience entries $\epsilon$ in natural language that encode reusable reasoning knowledge. We categorize experiences into two complementary types:
\begin{itemize}
    % 成功案例中提炼的有效推理策略和诊断原则
    \item \textbf{Golden Experience}: distilled from successful trajectories, capturing effective reasoning strategies, diagnostic principles, and validated decision patterns.
    % 失败案例中提炼的错误原因和应避免的模式
    \item \textbf{Warning Experience}: distilled from failed trajectories, summarizing failure causes, misleading cues, and recurring error patterns to avoid in future reasoning.
\end{itemize}

This dual categorization allows EMR to learn symmetrically from both successes and failures, preventing overfitting to correct outcomes while promoting robust error awareness. Formally:
\begin{equation}
\epsilon = \mathrm{Collect}(\tau, y \mid r_c, \pi_c),
\end{equation}
where $r_c$ denotes the collector role and $\pi_c$ the experience mining prompt. The collector compares the trajectory outcome with the ground-truth label $y$, identifies salient reasoning behaviors or failure modes, and abstracts them into experience entries for long-term accumulation.

\subsection{Experience Reuse} 
\label{subsection:reuse}
% EMR 通过在推理时动态重用积累的经验支持持续学习，包含分层经验库、更新机制和检索机制三个组件。
EMR supports continual evolution by systematically reusing accumulated experiences during inference, comprising three components: hierarchical experience library (\ref{subsubsection:hierarchical_experience_library}), update mechanism (\ref{subsubsection:experience_update_mechanism}), and experience retrieval (\ref{subsubsection:retrieval}).

\begin{figure}
    \centering
    \includegraphics[width=1\linewidth]{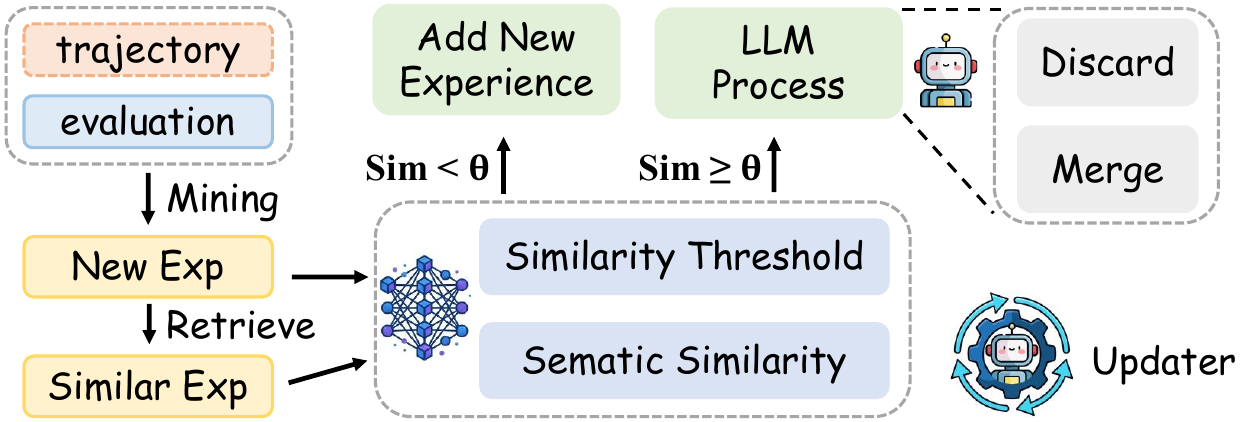}
    \caption{The process of updating the experience library.}
    \label{fig:updater}
\end{figure}

\subsubsection{Hierarchical Experience Library}
\label{subsubsection:hierarchical_experience_library}
% 多智能体经验系统需在不造成内存无序增长的前提下积累知识。EMR 采用分层经验库在不同抽象层次组织经验。
A key challenge in experience-driven multi-agent systems is accumulating knowledge without uncontrolled growth or sacrificed generalization. EMR adopts a hierarchical experience library $\mathcal{L}$ with three abstraction levels:
\begin{equation}
\mathcal{L} = \{\mathcal{L}^{\text{principle}}, \mathcal{L}^{\text{pattern}}, \mathcal{L}^{\text{case}}\}.
\end{equation}
At the highest level, $\mathcal{L}^{\text{principle}}$ stores abstract diagnostic rules and high-level reasoning principles. The intermediate level $\mathcal{L}^{\text{pattern}}$ captures reusable reasoning structures and decision templates. At the lowest level, $\mathcal{L}^{\text{case}}$ contains concrete, context-specific examples preserving detailed clinical reasoning traces. This hierarchical organization balances abstraction and specificity, enabling both compact storage and flexible reuse.

\subsubsection{Experience Update Mechanism}
\label{subsubsection:experience_update_mechanism}

Given a newly generated experience $\epsilon$, the updater agent determines whether and how to incorporate it into the library. The core objective is to preserve coherence and compactness while allowing novel experience to accumulate. Equivalent or redundant entries are discarded or merged rather than blindly added.

The update process consists of three steps. First, the updater retrieves potentially related experiences:
\begin{equation}
\mathcal{R} = \mathrm{Retrieve}(\mathcal{L}_t, \epsilon),
\end{equation}
where $\mathcal{L}_t$ denotes the library at step $t$.

Second, the updater estimates semantic relevance between $\epsilon$ and $\mathcal{R}$. Each experience is encoded into a semantic vector, and similarity is computed via cosine similarity:
\begin{equation}
\mathrm{Sim}(\epsilon, \mathcal{R}) = \max_{\epsilon_i \in \mathcal{R}}
\frac{\mathbf{e}(\epsilon) \cdot \mathbf{e}(\epsilon_i)}
{\lVert \mathbf{e}(\epsilon) \rVert , \lVert \mathbf{e}(\epsilon_i) \rVert},
\end{equation}
where $\mathbf{e}(\cdot)$ denotes the embedding function.

Finally, the library is updated according to:
\begin{equation}
\begin{aligned}
&\mathcal{L}_{t+1} = \\
&\quad
\begin{cases}
\mathcal{L}_t \cup \{\epsilon\}, \
\text{if } \mathrm{Sim}(\epsilon, \mathcal{R}) < \theta, \\[6pt]
(\mathcal{L}_t \setminus \mathcal{R}) \cup 
\mathrm{LLM\_Process}(\mathcal{R}, \epsilon),
 \text{otherwise}.
\end{cases}
\end{aligned}
\end{equation}
When similarity falls below $\theta$, the experience is considered novel and added directly. Otherwise, the LLM determines whether to merge them into a more compact representation or discard $\epsilon$ if it adds no reasoning value.

% 该机制结合轻量级相似度路由与 LLM 语义整合，实现可扩展的经验积累。
This mechanism enables scalable experience accumulation by combining lightweight similarity-based routing with LLM-based semantic consolidation, ensuring the library evolves in a structured and reusable manner.

\subsubsection{Experience Retrieval Mechanism}
\label{subsubsection:retrieval}
% EMR 将经验库视为外部知识源，在推理时动态查询以指导智能体决策。
Accumulated experience is only valuable if effectively reused during future reasoning. Rather than statically encoding experience into model parameters, EMR treats the experience library as an external, model-agnostic knowledge source dynamically queried at inference time.

Given a query context $c$, the retrieval mechanism selects relevant experiences:
\begin{equation}
\mathcal{E}_r = \mathrm{Retrieve}(\mathcal{L}, c, k),
\end{equation}
where $k$ denotes the number of retrieved entries. Retrieval proceeds hierarchically: the agent first identifies relevant high-level principles, then retrieves associated reasoning patterns, and finally accesses concrete cases grounding abstract knowledge in specific scenarios.

The retrieved experiences are injected into the planner, department, and summary agents as external guidance, shaping reasoning behavior without modifying model parameters. By conditioning reasoning on accumulated experience, EMR enables agents to benefit from prior successes, avoid previously observed failures, and generalize beyond their original training distribution.
% ==================== 主表格 ====================
\begin{table*}[ht]
\caption{Main results (Accuracy \%) on various medical benchmarks. The datasets include MedQA (USMLE), MedMC. (representing MedMCQA, Indian medical exams), PubMed. (representing PubMedQA, biomedical research questions), and various subsets of MMLU and MMLU-Pro, specifically: Ant. (Anatomy), Clin. (Clinical Knowledge), Coll. (College Medicine), Gen. (Medical Genetics), Hea. (Health), and Bio. (Biology). Bold indicates the best performance within each backbone model group.}
\label{tab:comparison_flattened}
\centering
\small
\renewcommand{\arraystretch}{1.4}
\setlength{\tabcolsep}{3pt}

\begin{tabularx}{0.95\textwidth}{l*{10}{Y}}
\toprule
\textbf{Method} & 
\textbf{MedQA} & 
\textbf{MedMC.} & 
\textbf{PubMed.} & 
\textbf{Ant.} & 
\textbf{Clin.} & 
\textbf{Coll.} & 
\textbf{Gen.} & 
\textbf{Hea.} & 
\textbf{Bio.} & 
\textbf{Avg.} \\ \midrule

% ================= Qwen3-8B Section =================
\rowcolor{gray!20} \multicolumn{11}{c}{\textbf{Qwen3-8B}} \\
Single-LLM & 70.2 & 66.5 & 72.1 & 68.8 & 71.4 & 69.5 & 73.2 & 60.1 & 65.8 & 68.6 \\
MedAgents  & 75.4 & 72.3 & 77.2 & 76.5 & 79.1 & 75.8 & 78.4 & 66.5 & 73.6 & 75.0 \\
MDAgents   & 77.8 & 75.6 & 80.5 & 78.9 & 81.5 & 77.2 & 82.1 & 69.5 & 76.8 & 77.8 \\
ExpeL & 76.5 & 73.8 & 79.2 & 77.5 & 80.1 & 76.8 & 80.5 & 68.2 & 75.1 & 76.4 \\
ReasoningBank & 77.2 & 74.5 & 79.8 & 78.2 & 80.8 & 77.5 & 81.2 & 68.8 & 75.8 & 77.1 \\
G-Memory & 77.8 & 75.1 & 80.2 & 78.8 & 81.4 & 77.8 & 81.8 & 69.2 & 76.3 & 77.6 \\
MDTeamGPT & 78.5 & 76.2 & 81.0 & 79.5 & 82.1 & 78.5 & 82.6 & 70.1 & 77.0 & 78.4 \\
EvolveR & 79.8 & 77.0 & 82.1 & 80.8 & 83.5 & 79.8 & 83.9 & 71.2 & 78.1 & 79.6 \\
\textbf{EMR (Ours)} & \textbf{81.0} & \textbf{78.8} & \textbf{83.5} & \textbf{82.2} & \textbf{85.4} & \textbf{80.9} & \textbf{84.7} & \textbf{73.1} & \textbf{80.5} & \textbf{81.1} \\ \midrule \midrule

% ================= GPT-4o Section =================
\rowcolor{gray!20} \multicolumn{11}{c}{\textbf{GPT-4o}} \\
Single-LLM & 86.5 & 83.1 & 82.4 & 85.9 & 88.1 & 87.2 & 89.5 & 79.2 & 83.4 & 85.0 \\
MedAgents  & 91.5 & 87.4 & 89.2 & 90.5 & 94.1 & 91.8 & 93.2 & 83.4 & 88.9 & 90.0 \\
MDAgents   & 93.8 & 90.5 & 91.6 & 92.8 & 96.2 & 93.5 & 95.8 & 85.6 & 91.4 & 92.4 \\
ExpeL & 87.4 & 84.1 & 83.5 & 86.8 & 89.2 & 88.1 & 90.3 & 80.1 & 84.5 & 86.0 \\
ReasoningBank & 88.6 & 85.5 & 84.8 & 88.1 & 90.5 & 89.2 & 91.6 & 81.3 & 85.8 & 87.2 \\
G-Memory & 89.3 & 84.9 & 85.2 & 88.5 & 90.9 & 89.6 & 92.0 & 81.8 & 86.2 & 87.6 \\
MDTeamGPT & 90.1 & 88.6 & 86.5 & 89.4 & 92.1 & 90.8 & 93.1 & 82.9 & 87.5 & 88.9 \\
EvolveR & 92.7 & 86.5 & 88.1 & 91.2 & 93.8 & 92.5 & 94.9 & 84.2 & 89.1 & 90.3 \\
\textbf{EMR (Ours)} & \textbf{96.8} & \textbf{89.2} & \textbf{93.4} & \textbf{94.7} & \textbf{97.5} & \textbf{95.4} & \textbf{98.2} & \textbf{86.8} & \textbf{93.1} & \textbf{93.9} \\ \bottomrule
\end{tabularx}
\end{table*}

\section{Experiments}

\subsection{Experimental Setup}

\noindent \textbf{Datasets.} We evaluate EMR on a diverse set of medical reasoning benchmarks, including MedQA~\cite{jin2021disease}, MedMCQA~\cite{pal2022medmcqa}, PubMedQA~\cite{jin2019pubmedqa}, MMLU~\cite{hendrycks2020measuring}, and MMLU-Pro~\cite{wang2024mmlu}.
These datasets cover a wide range of medical problem types, from clinical decision-making to complex medical reasoning, providing a comprehensive testing platform for evaluating medical multi-agent systems.

\noindent \textbf{Implementation.} We instantiate EMR with Qwen3-8B~\cite{qwen3technicalreport} and GPT-4o~\cite{hurst2024gpt} to examine robustness across model scales and architectures.
The numbers of domain experts for the question and options are set as: $m = 3, n = 2$. Analogous to training epochs, we set the number of data mining iterations to 5 in our implementation. More implementation details see Appendix~\ref{apppendix:Implementation Details}. All reported results are averaged over three independent runs.

\noindent \textbf{Baselines.} We compare EMR against both single-LLM and medical multi-agent baselines.
We use zero-shot prompting~\cite{kojima2022large} as a single-LLM baseline.
Multi-agent baselines include MedAgents~\cite{tang2024medagents} and MDAgents~\cite{kim2024mdagents}, which represent state-of-the-art static medical multi-agent systems.
Additionally, we compare against experience-based reasoning methods including ExpeL~\cite{zhao2024expel}, ReasoningBank~\cite{ouyang2025reasoningbank}, G-Memory~\cite{zhang2026g}, MDTeamGPT~\cite{chen2025mdteamgpt}, and EvolveR~\cite{wu2025evolver}.

\subsection{Main Results}

\textbf{Comparison with static multi-agent systems.}
Table~\ref{tab:comparison_flattened} shows that EMR consistently outperforms existing medical multi-agent systems across all LLMs and datasets. Under Qwen3-8B, EMR raises the average accuracy from 77.8 (MDAgents) to 81.1; under GPT-4o, it reaches 93.9. Unlike single-LLM baselines, EMR decomposes complex problems and enables structured collaboration. Compared to prior multi-agent methods, EMR introduces explicit experience mining and reuse, avoiding static role assignment and recurring failures.

\textbf{Comparison with experience-based methods.}
Among experience-based approaches, EvolveR achieves 92.7 (MedQA) and 86.5 (MedMCQA) via GRPO optimization, while MDTeamGPT reaches 88.6 on MedMCQA with flat consultation summaries. Others (ExpeL, ReasoningBank, G-Memory) lag due to limited abstraction or adaptation. EMR surpasses all, achieving 96.8 on MedQA (+4.1) and 89.2 on MedMCQA (+0.6), with larger gains on complex USMLE-style questions. On MMLU-Pro (Health \& Biology), EMR consistently beats all baselines across LLMs. Notably, EMR also boosts smaller LLMs like Qwen3-8B, closing the gap with larger models, while still enhancing strong LLMs such as GPT-4o.e-driven reasoning complements model capacity rather than replacing it.

\begin{table}[ht]
\caption{Ablation study on MedQA and MedMC datasets with Qwen3-8B. Exp: Experience, Pla: Planner, Dep: Department Agents.}
\centering
\small
\renewcommand{\arraystretch}{1.2}
\begin{tabular}{lcc}
\toprule
\textbf{Method} & \textbf{MedQA} & \textbf{MedMC} \\
& \textbf{Acc. (\%)} & \textbf{Acc. (\%)} \\
\midrule
Single-LLM & 70.2 & 66.5 \\
\midrule
\textbf{EMR (Full)} & \textbf{81.0} & \textbf{78.8} \\
\quad w/o Exp & 74.3 & 71.2 \\
\quad w/o Exp + Pla & 72.8 & 69.5 \\
\quad w/o Exp + Pla + Dep & 70.5 & 67.8 \\
\quad w/o Pla + Dep & 78.1 & 75.4 \\
\bottomrule
\end{tabular}
\label{tab:ablation_of_componets}
\end{table}

\section{Ablation Study}
\subsection{Ablation Study on EMR Components}
\label{subsection:ablation of components}
To assess the contribution of each EMR component, we conduct ablation experiments by progressively removing key modules (Table~\ref{tab:ablation_of_componets}). Removing experience mining and retrieval (w/o Exp) causes a 6.7\% drop, reducing EMR to the level of static systems like MedAgents. This shows that multi-agent collaboration alone is insufficient without experience accumulation. Among all components, experience yields the largest performance gain, underscoring its central role in enabling continuous improvement.

\begin{table}[ht]
\caption{Ablation study for the sub-components of the Experience module on MedQA and MedMC datasets with Qwen3-8B.}
\centering
\small
\renewcommand{\arraystretch}{1.2}
\begin{tabular}{lcc}
\toprule
\textbf{Method} & \textbf{MedQA} & \textbf{MedMC} \\
& \textbf{Acc. (\%)} & \textbf{Acc. (\%)} \\
\midrule
\textbf{EMR (Full)} & \textbf{81.0} & \textbf{78.8} \\
\midrule
\quad w/o Cases      & 79.8 & 77.4 \\
\quad w/o Patterns   & 78.5 & 75.9 \\
\quad w/o Principles & 76.4 & 73.5 \\
\bottomrule
\end{tabular}
\label{tab:ablation_experience_extended}
\end{table}

% 有趣的是，当移除规划员和部门代理，但保留经验信息（即不包含规划员和部门代理）时，EMR 仍然保持着强劲的性能，与完整模型相比仅下降了 2.9%。这表明，积累的经验可以通过直接指导汇总代理的推理，部分弥补角色分解的缺失。
Interestingly, when planner and department agents are removed but experience is retained (w/o Pla + Dep), EMR still maintains strong performance, with only a $2.9\%$ decrease compared to the full model. This suggests that accumulated experience can partially compensate for the absence of explicit role decomposition by directly guiding the summary agent’s reasoning.

\subsection{Ablation Study on Experience Hierarchy}
We analyze the impact of different experience hierarchy levels using the ablation results in Table~\ref{tab:ablation_experience_extended}. Removing case-level experiences leads to a modest 1.2\% drop, indicating that concrete historical cases offer useful but limited guidance. Ablating pattern-level experiences causes a larger 2.5\% degradation, suggesting that abstracted reasoning patterns are more important for generalization across questions. The most significant decline occurs when principle-level experiences are removed, with a substantial 4.6\% drop that brings EMR close to baseline performance. These results confirm that principle-level knowledge serves as the core transferable component for robust reasoning across diverse medical scenarios, while all three levels work complementarily to support overall system performance.

\section{Analysis}

\subsection{Experience Transferability}
\label{subsection:experience_transferability}
% 诸如 GPT-4o 和 DeepSeek-V3 等更强大的模型挖掘的经验，在被规模较小的模型（例如 Qwen3-8B）重用时，能够带来显著的性能提升，这反映了它们更丰富的医学知识、更可靠的推理模式以及更高质量的故障总结。值得注意的是，这种可迁移性是双向的：即使是强大的骨干模型也能从重用规模较小的模型挖掘的经验中获益，这些经验提供了多样化的诊断视角、替代推理启发式方法和互补的错误模式。总而言之，这些结果表明，在多属性自动误差（EMR）中，经验重用并非仅仅取决于模型容量，而是得益于异构推理行为的丰富性，从而构建出一个更稳健、更多样化的经验库。
Table~\ref{tab:med_scaling_advanced} shows that experiences mined by stronger models such as GPT-4o and DeepSeek-V3 \cite{deepseekai2024deepseekv3technicalreport} yield substantial gains when reused by smaller models (e.g., Qwen3-8B), reflecting their richer medical knowledge, more reliable reasoning patterns, and higher-quality failure summaries. Notably, this transferability is bidirectional: even strong LLMs benefit from reusing experiences mined by smaller models, which contribute diverse diagnostic perspectives, alternative reasoning heuristics, and complementary error patterns.

\begin{table}[ht]
\caption{The impact of experience from different models on other models.}
\centering
\small
\renewcommand{\arraystretch}{1.2}
\begin{tabular}{lccc}
\toprule
 & \textbf{w/o Exp} & \textbf{+ Qwen} & \textbf{+ GPT-4o} \\
\midrule
\multicolumn{4}{c}{\textbf{MedQA}} \\
\midrule
\textit{Qwen3} & 74.3 & 81.0 \gain{6.7} & 83.9 \gain{9.6} \\
\textit{DeepSeek} & 86.2 & 92.5 \gain{6.3} & 94.5 \gain{8.3} \\
\midrule
\multicolumn{4}{c}{\textbf{MedMCQA}} \\
\midrule
\textit{Qwen3} & 71.3 & 78.8 \gain{7.5} & 81.2 \gain{9.9} \\
\textit{DeepSeek} & 79.1 & 84.6 \gain{5.5} & 86.1 \gain{7.0} \\
\midrule
\multicolumn{4}{c}{\textbf{PubMedQA}} \\
\midrule
\textit{Qwen3} & 76.2 & 83.5 \gain{7.3} & 86.2 \gain{10.0} \\
\textit{DeepSeek} & 81.5 & 86.5 \gain{5.0} & 88.4 \gain{6.9} \\
\bottomrule
\end{tabular}
\label{tab:med_scaling_advanced}
\end{table}

\begin{table}[ht]
\caption{Cross-dataset experience analysis.}
\centering
\small
\renewcommand{\arraystretch}{1.2}
% 调整列间距以适应页面
\setlength{\tabcolsep}{12pt}
\begin{tabular}{lcc}
\toprule
\textbf{Source} & \textbf{MedQA} & \textbf{MedMCQA} \\
\midrule
\textit{MedQA} & \textbf{81.0} \gain{6.7} & 74.8 \gain{3.5} \\
\textit{MedMCQA} & 77.5 \gain{3.2} & \textbf{78.8} \gain{7.5} \\
\bottomrule
\end{tabular}
\label{tab:experience_cross_analysis}
\end{table}

% 表xx展示了经验可迁移性的跨数据集分析。虽然从同一数据集挖掘的经验在其对应的基准测试中表现最为显著，但从一个数据集提取的经验在迁移到不同的数据集时也能持续提升性能。这表明经验库捕获的是可通用的医学推理知识，而非特定于某个数据集的启发式方法。值得注意的是，积累的经验在具有不同问题风格和分布的数据集上仍然有效，这表明经验库可以作为医学多智能体推理的可迁移和可重用的知识资源。
Table~\ref{tab:experience_cross_analysis} presents a cross-dataset analysis of experience transferability. While experiences mined from the same dataset yield the strongest improvements on their corresponding benchmarks, experiences extracted from one dataset also consistently improve performance when transferred to different datasets. This indicates that the experience library captures generalizable medical reasoning knowledge rather than dataset-specific heuristics. Notably, the accumulated experiences remain effective across datasets with different question styles and distributions, demonstrating that the experience library serves as a transferable and reusable knowledge resource for medical multi-agent reasoning.

\subsection{Experience Analysis}
\label{subsection:experience_analysis}

% 通过对挖掘出的推理轨迹进行定性分析，我们
% 识别出医学多智能体推理中几个反复出现的失败模式。
% 依赖参数知识而非临床指南，是指智能体默认依赖内部记忆的事实，而不是明确地运用权威的诊断或治疗标准进行推理。
% 经常观察到医学概念混淆，即密切相关的疾病、症状或临床术语被错误地混淆，导致中间推理出现缺陷。
% 当智能体专注于表面相关性而未能识别疾病的潜在病理生理原因时，就会出现表面因果推理。
% 当多个智能体在缺乏充分交叉验证的情况下强化错误的假设时，就会出现多智能体协作中的集体幻觉，
% 这种集体幻觉会放大共同的错误，而不是纠正它们。
Through qualitative analysis of mined reasoning trajectories, we identify several recurring failure patterns in medical multi-agent reasoning.
\textbf{(1) Reliance on parametric knowledge rather than clinical guidelines} occurs when agents default to internally memorized facts instead of explicitly reasoning with authoritative diagnostic or treatment standards.
\textbf{(2) Medical concept confusion} is frequently observed, where closely related conditions, symptoms, or clinical terms are incorrectly conflated, leading to flawed intermediate reasoning.
\textbf{(3) Superficial causal reasoning} arises when agents focus on surface-level correlations while failing to identify the underlying pathophysiological cause of a condition.
Finally, \textbf{(4) collective hallucination in multi-agent collaboration} emerges when multiple agents reinforce an incorrect assumption without sufficient cross-verification, amplifying shared errors instead of correcting them. Examples see Appendix \ref{appendix:Experience_Examples}.

\section{Conclusion}

We presented EMR, a self-evolving medical multi-agent system driven by experience mining and reuse. By organizing golden and warning experiences into principle-, pattern-, and case-level knowledge, EMR progressively improves medical reasoning quality and robustness. Experiments across multiple medical benchmarks demonstrate consistent gains over strong multi-agent baselines, while ablation and transfer studies verify the effectiveness and reusability of the experience library. Overall, our results suggest that explicit experience modeling provides a scalable and interpretable framework for improving medical multi-agent systems and experience-driven collaborative LLM agents.

\section*{Limitations}

Despite the promising performance of EMR, several limitations remain.

First, EMR currently focuses on text-based medical question answering benchmarks and does not incorporate multimodal clinical information such as medical imaging, laboratory signals, or electronic health records. Real-world clinical decision-making often depends on heterogeneous data sources, and extending experience mining and reuse to multimodal settings remains an important direction for future work.

Second, the quality of the experience library is inherently dependent on the reasoning quality of the underlying LLMs. Although EMR can extract both golden and warning experiences, incorrect or hallucinated reasoning trajectories may still introduce noisy experiences into the library. While the similarity-based update mechanism partially mitigates this issue, more robust experience verification and filtering strategies are still needed.

Third, the current experience retrieval mechanism mainly relies on semantic similarity and hierarchical abstraction, which may not fully capture complex causal relationships between medical cases. As the experience library grows, retrieval efficiency and long-term maintenance may also become increasingly challenging.

Finally, although EMR improves reasoning performance on benchmark datasets, our experiments are conducted in controlled offline evaluation settings rather than real clinical environments. The system has not been validated for real-world medical deployment, and its outputs should not be interpreted as clinical advice. Future work should investigate human-in-the-loop evaluation, clinical safety constraints, and collaboration with healthcare professionals before practical adoption.

\bibliography{custom}

\appendix

\appendix

\section{Implementation Details}
\label{apppendix:Implementation Details}
For MedQA and MedMCQA, we split the datasets evenly, using half for experience mining and the other half for evaluation. For PubMedQA, 500 instances are used for experience mining and another 500 (official test set) for evaluation. MMLU (anatomy, clinical knowledge, college medicine, medical genetics) and MMLU-Pro (health, biology) are evenly split between experience mining and testing, with 200 instances per selected MMLU-Pro subset. We adopt bge-m3 \cite{bge-m3} as the semantic embedding model and set the similarity threshold to 0.85 (See Appendix \ref{app:theta} for more details). The number of retrieved experience is 3.
Statistically, the cost of experience mining is approximately \$5 per 100 samples (5 epochs), with a total runtime of about 2 hours, while inference takes around 45 seconds per sample.

\section{Sensitivity Analysis of Similarity Threshold $\theta$}
\label{app:theta}

The similarity threshold $\theta$ controls how strictly the retriever filters historical experiences during the update phase. A low threshold admits noisy or irrelevant experiences, while an excessively high threshold may leave the experience bank too sparse to provide meaningful guidance. To determine the optimal operating point, we sweep $\theta \in \{0.70, 0.80, 0.85, 0.90\}$ on the MedQA and MedMCQA validation sets using the Qwen3-8B backbone.

\begin{table}[h]
\caption{Sensitivity of the experience update mechanism to the similarity threshold $\theta$ and ablation on experience types (Qwen3-8B).}
\label{tab:app_theta}
\centering
\small
\begin{tabular}{lcc}
\toprule
\textbf{Setting} & \textbf{MedQA} & \textbf{MedMCQA} \\
\midrule
$\theta = 0.70$ & 80.1 & 77.9 \\
$\theta = 0.80$ & 80.7 & 78.5 \\
$\theta = 0.85$ (default) & 81.0 & 78.8 \\
$\theta = 0.90$ & 80.5 & 78.3 \\
\midrule
Golden-only & 78.4 & 76.2 \\
Golden + Warning (Ours) & 81.0 & 78.8 \\
\bottomrule
\end{tabular}
\end{table}

\textbf{Threshold sensitivity.} As shown in Table~\ref{tab:app_theta}, $\theta = 0.85$ yields the best performance on both benchmarks (81.0 on MedQA and 78.8 on MedMCQA). When $\theta$ drops to 0.70, accuracy falls by 0.9 and 0.9 points, respectively, indicating that overly permissive retrieval introduces distractive or low-quality experiences that mislead the reasoning process. Conversely, raising $\theta$ to 0.90 degrades performance by 0.5 and 0.5 points, suggesting that excessive filtering prunes useful cases and reduces coverage of the experience bank. The sweet spot at 0.85 strikes a balance between precision and recall of retrieved experiences.

\textbf{}{Ablation on experience types.} We further ablate the contribution of warning experiences. Using only golden (correct) experiences results in a substantial drop of 2.6 points on MedQA and 2.6 points on MedMCQA compared to our full setting. This confirms that warning experiences—capturing failed reasoning patterns and error-prone knowledge boundaries—are complementary to golden ones. They act as guardrails that help the model recognize and avoid previously encountered mistakes, thereby improving robustness on challenging medical questions.

\section{Prompts}

\subsection{Prompt for Medical Experience Collection from Multi-Agent Reasoning}

\begin{mdframed}[linewidth=0.8pt, roundcorner=5pt, backgroundcolor=gray!5]
\small \setlength{\parindent}{0pt}
You are a senior medical education expert specializing in mining reusable medical experience from multi-agent clinical reasoning trajectories.

\vspace{0.8em}
\textbf{Task Description}

Given a complete multi-agent medical reasoning trajectory, you are required to:
\begin{itemize}[leftmargin=*, noitemsep, topsep=0pt]
  \item Determine whether the final prediction is correct.
  \item Analyze the reasoning behaviors throughout the multi-agent collaboration process.
  \item Extract structured and reusable medical experience.
\end{itemize}

\vspace{0.8em}
\textbf{Reasoning Trajectory}

\{trajectory\_text\}

\vspace{0.8em}
\textbf{Evaluation Criteria}

\begin{itemize}[leftmargin=*, noitemsep, topsep=0pt]
  \item Ground-truth answer: \{golden\_answer\_idx\} (\{golden\_answer\})
  \item Model prediction: \{pred\_answer\}
  \item Trajectory status: \{"Successful" if is\_success else "Failed"\}
\end{itemize}

\vspace{0.8em}
\textbf{Experience Collection Requirements}

Based on the above trajectory, please extract the following type.

\vspace{0.8em}
\textbf{Experience Levels}

\textbf{Principle} \\
High-level diagnostic rules and clinical guidelines. Stable, abstract, and broadly applicable. Limited to 30--50 words.

\textbf{Pattern} \\
Reusable reasoning templates and decision-making structures describing how to think and analyze. Limited to 30--50 words.

\textbf{Case} \\
Concrete clinical scenario examples. \{"Successful" if is\_success else "Failure"\}-oriented analysis. Limited to 40--60 words.

\vspace{0.8em}
\textbf{Important Notes}

\begin{itemize}[leftmargin=*, noitemsep, topsep=0pt]
  \item Analyze the complete reasoning trajectory, including Planner, Department Agents, and Summary Agent.
  \item Pay attention to consensus formation and disagreement resolution.
  \item Extract abstract and reusable experience rather than restating the original question.
\end{itemize}

\vspace{0.8em}
\textbf{Output Format}

\{ \\
  "is\_correct": "\{str(is\_success).lower()\}", \\
  "principle": "...", \\
  "pattern": "...", \\
  "case": "..." \\
\}
\end{mdframed}

\subsection{Prompt for Planner Agents}

\begin{mdframed}[linewidth=0.8pt, roundcorner=5pt, backgroundcolor=gray!5]
\small \setlength{\parindent}{0pt}
You are a medical expert who specializes in categorizing medical scenarios into specific areas of medicine.
You can leverage accumulated reasoning patterns from past experiences to improve domain assignment and avoid previously observed errors. You need to complete the following steps:

\begin{enumerate}[leftmargin=*, noitemsep]
    \item \textbf{Carefully read the medical scenario presented in the question:}
\end{enumerate}

'''\{question\}'''

\textbf{Retrieved Relevant Experience}

The following experiences are retrieved from the experience library. Please consider them when assigning domains:

\textbf{Principles} \\
\{principles\_str\}

\textbf{Patterns} \\
\{patterns\_str\}

\textbf{Cases} \\
\{cases\_str\}

\vspace{0.5em}
\textit{\textbf{Note:}} If the above experiences are not relevant to the current question, please ignore them; if they are valuable, please incorporate them into your domain assignment.

\begin{enumerate}[leftmargin=*, noitemsep, start=2]
    \item \textbf{Based on the medical scenario, classify the question into \{num\_domains\} different subfields of medicine.}
    \item \textbf{You should output in exactly the same format as:}
\end{enumerate}

'''\{domain\_format\}'''

\vspace{0.5em}
\textbf{Output only the formatted result, no explanation needed.}
\end{mdframed}

\subsection{Prompt for Department Agent Medical Analysis}

\begin{mdframed}[linewidth=0.8pt, roundcorner=5pt, backgroundcolor=gray!5]
\small \setlength{\parindent}{0pt}
You are a senior \{department\} expert with extensive clinical experience and profound medical knowledge. Based on your professional background and the retrieved relevant experiences, please conduct an in-depth analysis of the following medical multiple-choice question.

\textbf{Question Information}

\textbf{Question:} \\
\{question\}

\textbf{Options:} \\
\{options\}

\textbf{Retrieved Relevant Experiences}

The following experiences, including successful strategies (Golden) and lessons from failures (Warning), have been retrieved from the experience library. Please incorporate them into your analysis:

\{principles\_str\}

\{patterns\_str\}

\{cases\_str\}

\textit{\textbf{Note:}} \\
- \textbf{Golden} experiences demonstrate correct reasoning paths and verification strategies. \\
- \textbf{Warning} experiences highlight common error patterns and pitfalls to avoid. \\
- Combine both perspectives: leverage successful methods while proactively avoiding known errors.

\vspace{0.5em}
\textbf{Please use professional and rigorous medical language, demonstrating both your expertise and clinical reasoning process.}
\end{mdframed}

\subsection{Prompt for Summary Agent}

\begin{mdframed}[linewidth=0.8pt, roundcorner=5pt, backgroundcolor=gray!5]
\small \setlength{\parindent}{0pt}
You are a senior medical diagnostic expert specializing in integrating multidisciplinary opinions to make a final judgment. Based on the analysis from various department experts and the retrieved relevant experiences, please provide a comprehensive assessment for this medical multiple-choice question.

\textbf{Question Information}

\textbf{Question:} \\
\{question\}

\textbf{Options:} \\
\{options\_str\}

\textbf{Departmental Expert Analyses}

\{all\_analysis\}

\textbf{Retrieved Relevant Experiences}

The following experiences, including successful strategies (Golden) and lessons from failures (Warning), have been retrieved from the experience library. Please refer to these during your synthesis:

\{principles\_str\}

\{patterns\_str\}

\{cases\_str\}

\textit{\textbf{Note:}} \\
- \textbf{Golden} experiences demonstrate correct reasoning paths and verification strategies. \\
- \textbf{Warning} experiences highlight common error patterns and pitfalls to avoid. \\
- Leverage successful methods while proactively avoiding known errors mentioned in Warning experiences.

\textbf{Comprehensive Analysis Task}

Please complete the following steps:

\begin{enumerate}[leftmargin=*, noitemsep]    
    \item \textbf{Synthesized Judgment:} Weigh departmental opinions according to their professional relevance. Identify diagnostic traps or common errors (referencing Warning experiences) and apply the correct reasoning patterns from Golden experiences.
    
    \item \textbf{Final Answer:} Clearly state the correct option (must be one of \{list(options.keys())\}), summarize the core rationale for your choice, and explain how to prevent common mistakes related to this problem.
\end{enumerate}

\textbf{Output Format}

Please output in the following format:

\textbf{[Comprehensive Analysis]} \\
(Provide a detailed comprehensive analysis, including opinion consolidation, evidence analysis, and the judgment process.)

\textbf{[Final Answer]} \\
(Output only the option letter, e.g., A)
\end{mdframed}

\section{Case Study}

\begin{mdframed}[linewidth=0.8pt, roundcorner=5pt, backgroundcolor=gray!5]
\small \setlength{\parindent}{0pt}

\textbf{Question:} \\
A junior orthopaedic surgery resident is completing a carpal tunnel repair with the department chairman as the attending physician. During the case, the resident inadvertently cuts a flexor tendon. The tendon is repaired without complication. The attending tells the resident that the patient will do fine, and there is no need to report this minor complication that will not harm the patient, as he does not want to make the patient worry unnecessarily. He tells the resident to leave this complication out of the operative report. Which of the following is the correct next action for the resident to take?

\vspace{0.5em}
(A) Disclose the error to the patient and put it in the operative report \\
(B) Tell the attending that he cannot fail to disclose this mistake \\
(C) Report the physician to the ethics committee \\
(D) Refuse to dictate the operative report

\vspace{0.8em}
\textbf{Experts:} Medical Ethics, Orthopaedic Surgery, Medical Documentation

\vspace{0.8em}
\textbf{Analyses from various experts:}

\vspace{0.3em}
\textbf{Medical Ethics Expert:} Resident physicians should insist on recording complications in the surgical record and disclosing them to the patient. This is a direct reflection of upholding \textcolor{blue}{patient autonomy}, ensuring \textcolor{blue}{medical transparency}, and fulfilling professional responsibility. If superiors object, resident physicians should still insist and, if necessary, seek support from the hospital ethics committee.

\vspace{0.3em}
\textbf{Orthopaedic Surgery Expert:} As a senior orthopedic specialist, I emphasize that the growth of resident physicians includes learning to \textcolor{blue}{uphold professional standards under pressure}. In this case, despite pressure from superiors, the correct course of action was to honestly record and inform the patient. This protects both the patient and the physician's own \textcolor{blue}{professional integrity}.

\vspace{0.3em}
\textbf{Medical Documentation Expert:} From the perspective of a seasoned medical documentation expert, the \textcolor{blue}{completeness of medical records} is non-negotiable; any intraoperative event must be recorded truthfully. Furthermore, honesty with the patient is central to the \textcolor{blue}{doctor-patient relationship}. Faced with an erroneous instruction from superiors, the correct action is to uphold professional ethics.

\vspace{0.8em}
\textbf{Summary:} \\
\textcolor{magenta}{<<Experience Retrieved: When ethical obligations conflict with directives from superiors, one must distinguish between 'process steps' and 'final actions'.>>} \\
... \textcolor{blue}{Option B} is a reasonable "next step" because the superior has just given instructions, and the resident should first express their differing opinions and try to \textcolor{blue}{persuade the superior} to comply with professional ethics. If the superior still insists, the resident should record and disclose (A), but that is a second step.

\end{mdframed}

% \section{Future Work}
% Future work will enhance EMR through multimodal experience mining and federated learning, allowing agents to integrate visual and physiological data (e.g., medical imaging, ECG) into cross-modal “Golden Experiences” that connect biomarkers with diagnostic principles. At the same time, localized EMR instances can share only desensitized high-level principles and patterns across institutions, enabling learning from diverse patient populations while supporting complex clinical scenarios and preserving privacy.

\section{Experience Examples}
\label{appendix:Experience_Examples}
Below is a detailed example of the experience.

\begin{itemize}
  \item \textbf{Parametric Over-reliance.}  
  An agent recommends immediate PCI for a STEMI patient based on the general belief that ``PCI is superior,'' while neglecting a critical guideline constraint: when the expected delay to PCI exceeds 120 minutes, fibrinolysis should be the preferred reperfusion strategy.

  \item \textbf{Medical Concept Confusion.}  
  In a legal--medical reasoning task, agents conflate the regulatory definitions of \emph{hospital infection outbreak} and \emph{hospital infection prevalence}, resulting in an incorrect judgment of the mandatory reporting timeframe (12 hours vs.\ 24 hours).

  \item \textbf{Superficial Causal Reasoning.}  
  When confronted with a limb trauma case presenting numbness and coolness, agents focus on surface correlations and diagnose a nerve injury, failing to identify the underlying pathophysiological cause: acute arterial occlusion, in which neurological symptoms are secondary to ischemia.

  \item \textbf{Collective Hallucination.}  
  During a pharmacology discussion, multiple agents correctly recognize that 5-FU inhibits thymidylate synthase, but reinforce each other's false assumption that its structure mimics thymine (T), collectively overlooking the fact that 5-FU is a uracil (U) analogue.
\end{itemize}

\section{Ablation Study on the Number of Agents}
\label{subsection:ablation of agent number}
% 本研究探讨了智能体数量如何影响多智能体自动机（EMR）的性能和经验积累。其动机有二：首先，了解增加智能体数量是否能提升医学推理性能；其次，考察智能体多样性如何影响挖掘经验的多样性和规模。由于EMR依赖于协同推理和从智能体轨迹中挖掘经验，因此智能体数量直接决定了推理过程中可用的分析视角范围。
This study investigates how the number of agents affects the performance and experience accumulation of EMR. The motivation is twofold: first, to understand whether increasing the number of agents leads to better medical reasoning performance; and second, to examine how agent multiplicity influences the diversity and scale of mined experience. Since EMR relies on collaborative reasoning and experience mining from agent trajectories, the number of agents directly controls the breadth of analytical perspectives available during inference.

\begin{figure}[ht]
    \centering
    \includegraphics[width=\linewidth]{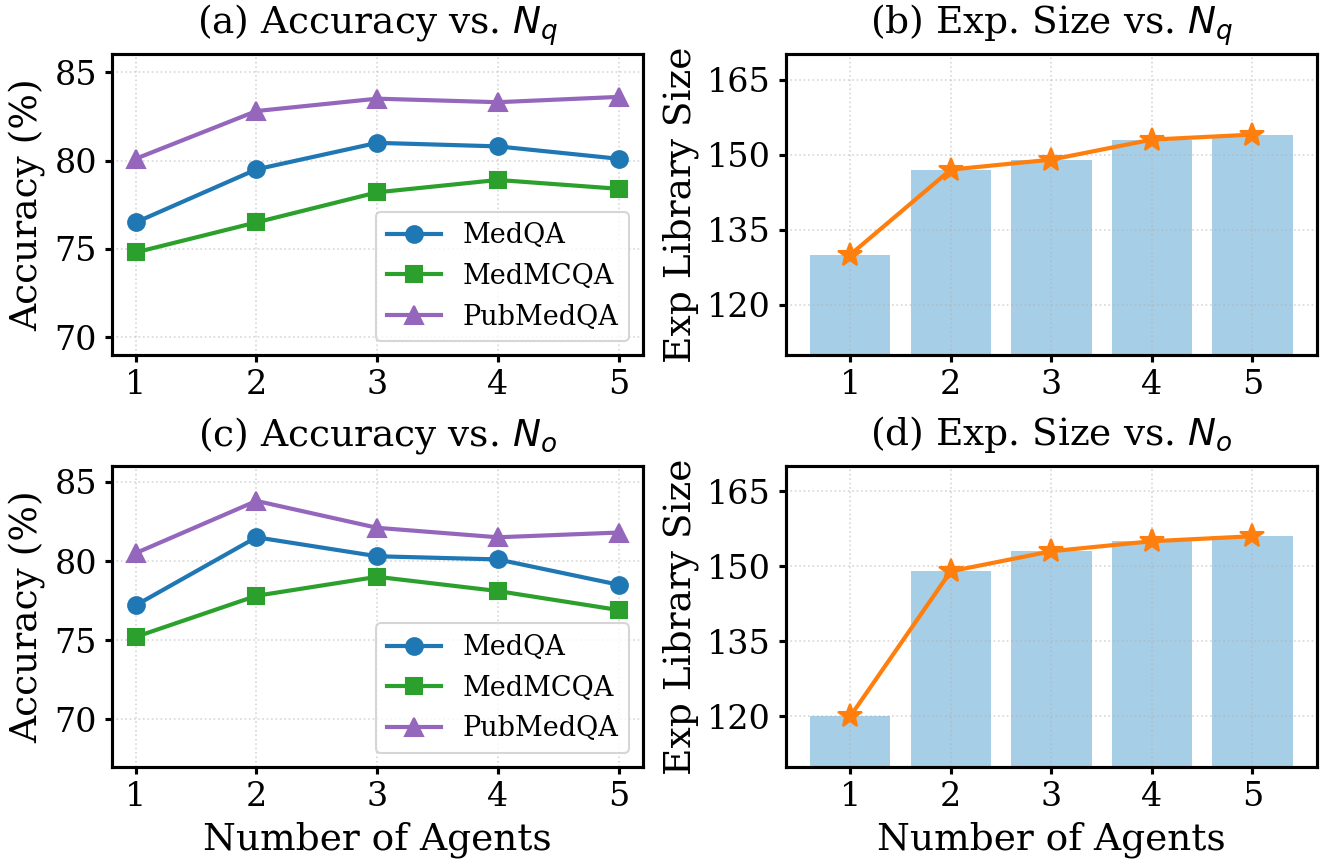}
    \caption{The impact of the number of agents on accuracy and experience library size. The ablation experiments are conducted on the Qwen3-8B model and the MedQA dataset.}
    \label{fig:ablation_of_agents}
\end{figure}

% 如图~\ref{fig:agent number ablation}所示，增加智能体的数量通常会提高所有基准测试的准确率，尤其是在智能体数量从少量增加到中等规模时。这一趋势表明，额外的智能体引入了互补的医学视角，从而能够更全面地探索诊断假设并减少推理盲点。然而，随着智能体数量的持续增长，性能提升逐渐趋于饱和，表明一旦达到足够的分析覆盖范围，收益就会递减。
As shown in Figure~\ref{fig:ablation_of_agents}, increasing the number of agents generally improves accuracy across all benchmarks, particularly when moving from a small number of agents to a moderate scale. This trend suggests that additional agents introduce complementary medical perspectives, enabling more comprehensive exploration of diagnostic hypotheses and reducing reasoning blind spots. However, performance gains gradually saturate as the number of agents continues to grow, indicating diminishing returns once sufficient analytical coverage is achieved. Beyond performance, increasing the number of agents substantially enlarges the experience library by enriching the diversity of reasoning trajectories, particularly when scaling from a small number of agents.

\section{Ablation Study on Retrieved Experience Size}
\begin{figure}
    \centering
    \includegraphics[width=1\linewidth]{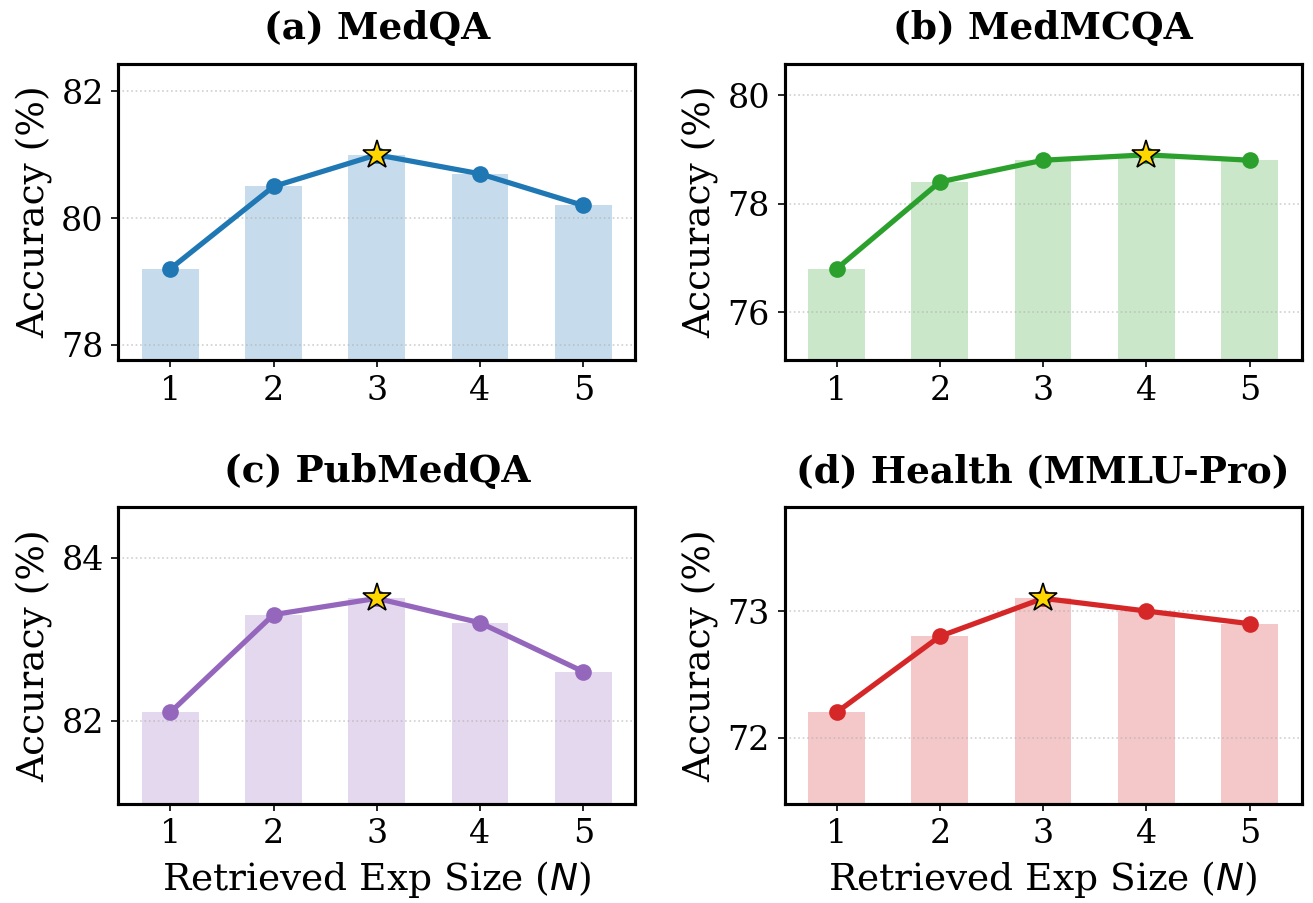}
    \caption{The impact of the number of retrieved experiences on performance. The ablation experiments are conducted using the Qwen3-8B model on the MedQA, MedMCQA, PubMedQA, and Health (MMLU-Pro) datasets.}
    \label{fig:ablation_of_exps}
\end{figure}

As shown in Figure~\ref{fig:ablation_of_exps}, increasing the number of retrieved experience blocks improves performance in the early stage, with the most significant gains observed when $N$ increases from 1 to 3. This suggests that a small set of diverse and highly relevant experiences is sufficient to effectively guide multi-agent reasoning by reusing successful diagnostic strategies and avoiding previously identified failure patterns.

Notably, the diminishing performance gains beyond this range indicate that EMR benefits more from the quality and abstraction level of reused experience than from sheer quantity. Since experiences in EMR are distilled into reusable principles and patterns, a limited number of experience blocks can already provide strong guidance, reflecting an efficient form of experience reuse rather than simple memory retrieval. From a clinical perspective, this trend aligns with human medical reasoning, where a few representative guidelines or prior cases often suffice to inform diagnosis, while excessive prior information may introduce redundancy or distraction. These findings further suggest that EMR achieves self-evolving not by increasing inference-time context indiscriminately, but by progressively mining and reusing compact, high-level experience that remains effective even under limited retrieval budgets.

\section{Effect of Experience Library Size}
\label{subsection:exp_scaling}
We evaluate EMR over multiple experience accumulation epochs, where newly collected experiences from agent reasoning trajectories are incrementally merged into the experience library. We track both the growth of the experience library and the corresponding test accuracy to examine how medical reasoning performance evolves with accumulated clinical experience.

\begin{figure}
    \centering
    \includegraphics[width=\linewidth]{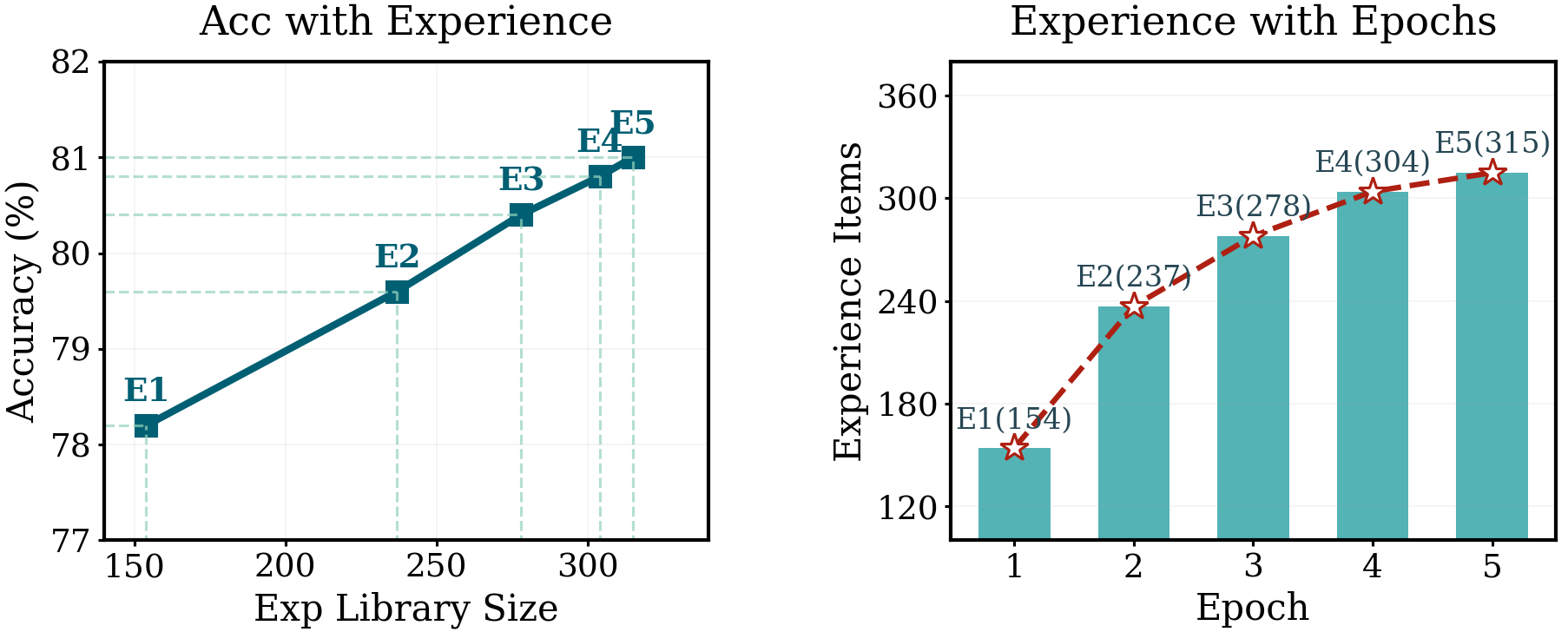}
    \caption{The size of the experience library varies with the number of mining epochs. The analysis experiments are conducted on the Qwen3-8B model and the MedQA dataset.}
    \label{fig:exp_size}
\end{figure}

As shown in Figure~\ref{fig:exp_size}, test accuracy consistently improves as the experience library expands across epochs. Performance gains are particularly pronounced in the early epochs, indicating that newly mined experiences effectively capture high-value clinical knowledge, such as common diagnostic reasoning patterns and frequently observed failure modes, which can be immediately reused to guide subsequent medical decision making.
As the number of epochs increases, the discovery rate of novel and informative clinical experiences gradually decreases, leading to a slower growth of the experience library. Correspondingly, performance improvements become more moderate and eventually stabilize. This saturation effect suggests that EMR progressively distills the most salient medical experience embedded in agent trajectories, after which additional accumulation primarily yields redundant or marginal clinical guidance.

\section{Computational Cost \& Efficiency Analysis}

As shown in Table~\ref{tab:efficiency_comparison}, both the token consumption and inference latency of EMR remain fully practical, while consistently outperforming existing medical multi-agent baselines in efficiency. Our multi-agent architecture is intentionally designed to be lightweight and efficient. Although inspired by MedAgents, EMR removes the expensive multi-round discussion mechanism commonly used for consensus-reaching, and instead introduces an experience-based conflict resolution strategy. During the entire collaborative reasoning process, the experience module consumes only around 300 additional tokens and requires approximately 2 seconds for retrieval.

In contrast, RECONCILE and MDAgents rely on more complex collaborative pipelines involving iterative reporting or multi-round discussions, which inevitably result in substantially higher token usage and longer inference time. These results demonstrate that explicit experience reuse can improve reasoning quality without introducing excessive computational overhead.

\begin{table*}[h]
\centering
\caption{Efficiency comparison of different methods.}
\label{tab:efficiency_comparison}
\small
\renewcommand{\arraystretch}{1.2}
\begin{tabular}{lcc}
\toprule
\textbf{Method} & \textbf{Avg. Tokens / Sample} & \textbf{Avg. Time / Sample (s)} \\
\midrule
Single-LLM (CoT) & 200 & 5 \\
MedAgents~\cite{tang2024medagents} & 1500 & 18 \\
RECONCILE~\cite{chen2024reconcile} & 2500 & 33 \\
MDAgents~\cite{kim2024mdagents} & 1900 & 23 \\
\textbf{EMR (Ours)} & \textbf{1200} & \textbf{15} \\
\bottomrule
\end{tabular}
\end{table*}

\section{Generalizability of EMR}

To further evaluate the generalizability of the proposed experience mechanism, we integrate EMR into existing medical multi-agent frameworks. To our knowledge, this is the first attempt to incorporate an explicit experience mining and reuse mechanism into medical multi-agent systems. Importantly, EMR is not restricted to a specific architecture and can be naturally extended to any multi-agent framework capable of collecting reasoning trajectories.

As shown in Table~\ref{tab:generalizability_EMR}, integrating EMR consistently improves the performance of both MDAgents and RECONCILE on MedQA and MedMCQA under the GPT-4o backbone. Specifically, MDAgents improves from 93.8 to 96.5 on MedQA and from 90.5 to 92.7 on MedMCQA after incorporating EMR. Similarly, RECONCILE achieves gains of 2.7 and 2.4 points, respectively.

These results suggest that the benefits of experience mining and reuse are architecture-agnostic and can serve as a general enhancement strategy for collaborative medical reasoning systems.

\begin{table}[h]
\centering
\caption{Generalizability of EMR under the GPT-4o backbone.}
\label{tab:generalizability_EMR}
\small
\renewcommand{\arraystretch}{1.2}
\begin{tabular}{lcc}
\toprule
\textbf{Method} & \textbf{MedQA} & \textbf{MedMCQA} \\
\midrule
MDAgents~\cite{kim2024mdagents} & 93.8 & 90.5 \\
MDAgents (w/ EMR) & \textbf{96.5} & \textbf{92.7} \\
\midrule
RECONCILE~\cite{chen2024reconcile} & 92.1 & 88.2 \\
RECONCILE (w/ EMR) & \textbf{94.8} & \textbf{90.6} \\
\bottomrule
\end{tabular}
\end{table}

\section{Comparison with Prior Experience-Based Methods}

As shown in Table~\ref{tab:comparison_experience_methods}, the hierarchical experience library in EMR differs fundamentally from prior experience-based methods in both representation and retrieval strategy, enabling a unified framework that balances generalization and specificity.

Existing approaches typically rely on storing raw reasoning trajectories or flat experience summaries, which often suffer from limited abstraction ability and weak transferability across tasks. In contrast, EMR organizes experience hierarchically into principles, patterns, and cases. Principle-level experiences capture abstract and transferable diagnostic knowledge, improving generalization across diverse medical reasoning scenarios. Pattern-level experiences encode reusable reasoning structures, while case-level experiences preserve detailed clinical contexts that directly support fine-grained reasoning.

Furthermore, prior methods generally retrieve only a few similar instances in a flat manner. EMR instead adopts hierarchical retrieval, progressively retrieving high-level principles, reasoning patterns, and representative cases, thereby enabling both broad reasoning guidance and detailed clinical grounding during inference.

\begin{table*}[h]
\centering
\caption{Comparison with prior experience-based methods.}
\label{tab:comparison_experience_methods}
\small
\renewcommand{\arraystretch}{1.2}
\begin{tabular}{p{3cm}p{5cm}p{5cm}}
\toprule
\textbf{Aspect} & \textbf{Prior Methods} & \textbf{EMR (Ours)} \\
\midrule
Representation & Raw trajectories / flat summaries & Hierarchical (principles, patterns, cases) \\
Generalization & Limited & Strong (via principles) \\
Detail Learning & Indirect & Direct (via cases) \\
Retrieval & Few instances & Hierarchical retrieval \\
\bottomrule
\end{tabular}
\end{table*}

\section{Sensitivity to Experience Library Size and Quality}

As shown in Table~\ref{tab:sensitivity_library_quality}, EMR remains robust under both limited and noisy experience settings, consistently outperforming standard multi-agent systems even when the experience library is significantly constrained or partially corrupted.

When only 20\% of the original experience library is retained, EMR still achieves 76.8 on MedQA and 75.4 on MedMCQA, substantially exceeding the performance of the system without experience. Even under the extreme setting where only 10\% of experiences are available, EMR preserves strong reasoning capability, indicating that a small number of high-level principles and reusable reasoning patterns already provide substantial guidance for medical decision-making.

We further evaluate robustness under noisy experience conditions by injecting irrelevant or low-quality experiences into the library. Although performance gradually declines as noise increases, EMR remains relatively stable, achieving 79.3 under 20\% noise and 78.4 under 40\% noise on MedQA. This robustness mainly stems from the hierarchical organization and experience merging mechanism, which mitigate the influence of noisy or redundant experiences through semantic consolidation and abstraction.

These findings suggest that EMR does not rely solely on large-scale memory accumulation. Instead, the principle-level abstraction and hierarchical refinement mechanisms enable the system to preserve strong reasoning performance even under sparse or imperfect experience conditions.

\begin{table}[h]
\centering
\caption{Sensitivity to experience library size and quality using the Qwen3-8B backbone.}
\label{tab:sensitivity_library_quality}
\small
\renewcommand{\arraystretch}{1.2}
\begin{tabular}{lcc}
\toprule
\textbf{Setting} & \textbf{MedQA} & \textbf{MedMCQA} \\
\midrule
\textbf{Full EMR} & \textbf{81.0} & \textbf{78.8} \\
Limited Exp (20\%) & 76.8 & 75.4 \\
Limited Exp (10\%) & 76.3 & 74.7 \\
Noisy Exp (+20\% noise) & 79.3 & 76.9 \\
Noisy Exp (+40\% noise) & 78.4 & 75.0 \\
w/o Experience & 75.3 & 72.1 \\
\bottomrule
\end{tabular}
\end{table}

\section{Potential Risks}

Although EMR is designed as a research framework for improving medical multi-agent reasoning, several potential risks should be acknowledged.

First, EMR may generate incorrect or misleading medical reasoning due to hallucinations, incomplete knowledge, or flawed collaboration among agents. Since the experience library accumulates knowledge from model-generated trajectories, erroneous reasoning patterns may propagate across future inferences if not properly filtered.

Second, the reuse of historical experiences may introduce bias amplification. Experiences mined from benchmark datasets or particular LLMs may overrepresent certain medical reasoning styles, disease distributions, or clinical assumptions, potentially reducing robustness when applied to diverse real-world populations or uncommon clinical cases.

Third, EMR may increase users’ overreliance on AI-generated medical suggestions. Although the system improves reasoning consistency, it is not a substitute for licensed medical professionals. Incorrect outputs in high-stakes clinical settings could lead to harmful medical decisions if used without expert supervision.

Fourth, experience storage and retrieval mechanisms may raise privacy and security concerns in practical deployments involving real patient data. While our experiments only use publicly available benchmark datasets, extending EMR to real-world clinical environments would require strict compliance with medical data protection regulations and careful anonymization of stored experiences.

Finally, the multi-agent collaboration process substantially increases computational cost and inference latency compared to single-agent systems, which may limit scalability and accessibility in resource-constrained healthcare environments.

\end{document}